\documentclass[runningheads]{llncs}
\usepackage{graphicx}

\usepackage{tikz}
\usepackage{comment}
\usepackage{amsmath,amssymb,amsfonts}
\usepackage{color}
\usepackage{balance}
\usepackage{bbm}
\usepackage{booktabs}
\usepackage{dsfont}
\usepackage{epsfig}
\usepackage{pifont}
\usepackage{graphicx}
\usepackage{multirow}
\usepackage[numbers,sort,compress,sectionbib]{natbib}
\usepackage{subcaption}
\usepackage{times}
\usepackage{url}
\usepackage[bottom]{footmisc}
\usepackage{comment}

\usepackage[accsupp]{axessibility}  %

\usepackage[colorlinks=true]{hyperref}%

\renewcommand{\refname}{References}
\makeatletter
\renewcommand{\bibsection}{%
   \section*{\refname%
            \@mkboth{\MakeUppercase{\refname}}{\MakeUppercase{\refname}}%
   }
}
\makeatother

\begin{document}
\pagestyle{headings}
\mainmatter
\def\ECCVSubNumber{1549}  %

\title{AlignSDF: Pose-Aligned Signed Distance Fields \\for Hand-Object Reconstruction}

\titlerunning{AlignSDF}
\author{Zerui Chen\inst{1} \and
Yana Hasson\inst{1,2} \and
Cordelia Schmid\inst{1} \and 
Ivan Laptev\inst{1}}
\authorrunning{Chen et al.}
\institute{Inria, \'Ecole normale sup\'erieure, CNRS, PSL Research Univ., 75005 Paris, France \and
Now at Deepmind \\
\email{\textbf{firstname.lastname@inria.fr}}\\
\url{https://zerchen.github.io/projects/alignsdf.html}}

\maketitle
\begin{abstract}
Recent work achieved impressive progress towards joint reconstruction of hands and manipulated objects from monocular color images. Existing methods focus on two alternative representations in terms of either parametric meshes or signed distance fields (SDFs). On one side, parametric models can benefit from prior knowledge at the cost of limited shape deformations and mesh resolutions. Mesh models, hence, may fail to precisely reconstruct details such as contact surfaces of hands and objects. SDF-based methods, on the other side, can represent arbitrary details but are lacking explicit priors. In this work we aim to improve SDF models using priors provided by parametric representations. In particular, we propose a joint learning framework that disentangles the pose and the shape. We obtain hand and object poses from parametric models and use them to align SDFs in 3D space. We show that such aligned SDFs better focus on reconstructing shape details and improve reconstruction accuracy both for hands and objects. We evaluate our method and demonstrate significant improvements over the state of the art on the challenging ObMan and DexYCB benchmarks.

\end{abstract}
\section{Introduction}
Reconstruction of hands and objects from visual data holds a promise to unlock wide\-spread applications in virtual reality, robotic manipulation and human-computer interaction. With the advent of deep learning, we have witnessed a large progress towards 3D reconstruction of hands~\cite{moon2020deephandmesh,zhou2020monocular,spurr2021self,boukhayma20193d,iqbal2018hand,cai20203d} and objects~\cite{choy20163d,groueix2018papier,wang2018pixel2mesh,peng2021shape}. Joint reconstruction of hands and manipulated objects, as well as detailed modeling of hand-object interactions, however, remains less explored and poses additional challenges. %

Some of the previous works explore 3D cues and perform reconstruction from multi-view images~\cite{chen2021mvhm}, depth maps~\cite{supanvcivc2018depth,zhang2021single,baek2018augmented} or point clouds~\cite{chen2018shpr}.
Here, we focus on a more challenging but also more practical setup and reconstruct hands and objects jointly from monocular RGB images. 
Existing methods in this setting can be generally classified as the ones using parametric mesh models~\cite{MANO:SIGGRAPHASIA:2017,hasson2019learning,hasson2020leveraging,qian2020html,yang2021cpf} and methods based on implicit representations~\cite{park2019deepsdf,mescheder2019occupancy,karunratanakul2020grasping,chen2019learning}.

\begin{figure*}[t]
  \centering
  \input{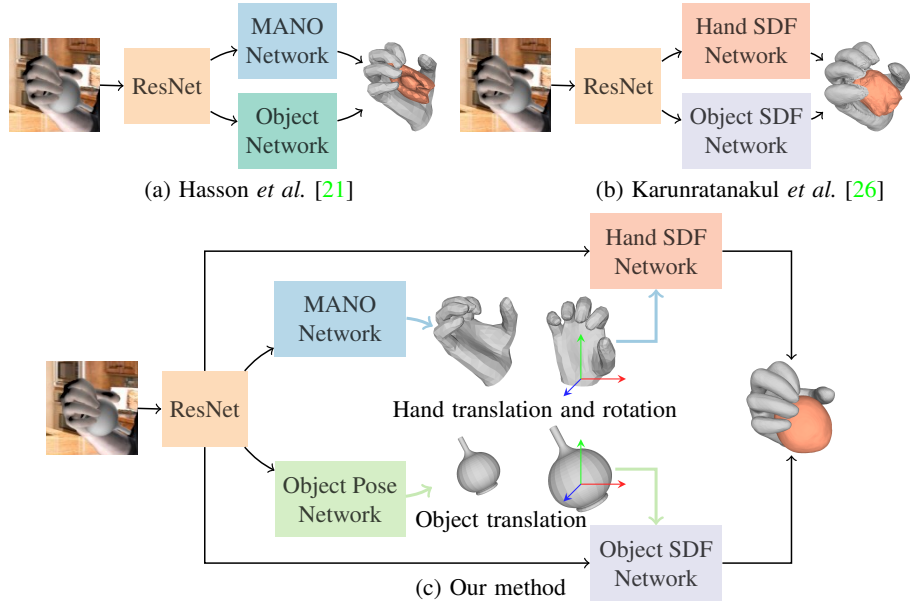}
  \caption{
  Previous work on hand-object reconstruction use either (a) parametric shape models or (b) implicit 3D representations. 
  Our proposed method (c) extends SDFs with prior knowledge on hand and object poses obtained via parametric models and can produce detailed meshes for hands and manipulated objects from monocular RGB images.}
  \label{motivation}
\end{figure*}

Methods from the first category~\cite{hasson2019learning,hasson2020leveraging,yang2021cpf} often
build on MANO~\cite{MANO:SIGGRAPHASIA:2017}, a popular parametric hand model, see Figure~\ref{motivation}(a). 
Since MANO is derived from 3D scans of real human hands and encodes strong prior shape knowledge, such methods typically provide anthropomorphically valid hand meshes. However, the resolution of parametric meshes is limited, making them hard to recover detailed interactions. Also, reconstructing 3D objects remains a big challenge. Hasson~\emph{et al.}~\cite{hasson2019learning} propose to use AtlasNet~\cite{groueix2018papier} to reconstruct 3D objects. However, their method can only reconstruct simple objects, and the reconstruction accuracy remains limited. To improve reconstruction, several methods~\cite{hasson2020leveraging,yang2021cpf,tekin2019h+} make a restricting assumption that the ground-truth 3D object model is available at test time and only predict the 6D pose of the object.

Recently, neural implicit representations have shown promising results for object reconstruction~\cite{park2019deepsdf}. Following this direction, Karunratanakul~\emph{et al.}~\cite{karunratanakul2020grasping} propose to represent hands and objects in a unified signed distance field (SDF) and show the potential to model hand-object interactions, see Figure~\ref{motivation}(b).  
We here adopt SDF and argue that such implicit representations may benefit from explicit prior knowledge about the pose of hands and objects.

For more accurate reconstruction of 
hands and manipulated objects, we attempt to combine the advantages of the parametric models and SDFs. Along this direction, previous works~\cite{deng2020nasa,saito2021scanimate,chen2021snarf,karunratanakul2021skeleton} attempt to leverage parametric models to learn SDFs from 3D poses or raw scans. In our work, we address a different and more challenging setup of reconstructing hands and objects from monocular RGB images. We hence propose a new pose-normalized SDF framework suited for our task. 

Scene geometry depends both on the shape and the global pose of underlying objects.
While the pose generally affects all object points with low-parametric transformations (e.g., translation and rotation), it is a common practice to separate the pose and the shape parameters of the model~\cite{cootes1995active,MANO:SIGGRAPHASIA:2017}. We, hence, propose to disentangle the learning of pose and shape for both hands and objects. As shown in Figure~\ref{motivation}(c), for the hand, we first estimate its MANO parameters and then learn hand SDF in a canonical frame normalized with respect to the rotation and trsnslation of the hand wrist.
Similarly, for objects, we estimate their translation and learn object SDF in a translation-normalized canonical frame.
By normalizing out the pose, we simplify the task of SDF learning which can focus on estimating the shape disregarding the global rotation and translation transformations. 
In our framework, the MANO network and the object pose network are responsible for solving the pose, and SDF networks focus on learning the geometry of the hand and the object under their canonical poses.

To validate the effectiveness of our approach, we conduct extensive experiments on two challenging benchmarks: ObMan~\cite{hasson2019learning} and DexYCB~\cite{chao2021dexycb}. ObMan is a synthetic dataset and contains a wide range of objects and grasp types. DexYCB is currently the largest real dataset for capturing hands and manipulated objects. We experimentally demonstrate that our approach outperforms state-of-the-art methods by a significant margin on both benchmarks. Our contributions can be summarized as follows:

$\bullet$ We propose to combine the advantages of parametric mesh models and SDFs and present a joint learning framework for 3D reconstruction of hands and objects.

$\bullet$ To effectively incorporate prior knowledge into SDFs learning, we propose to disentangle the pose learning from the shape learning for this task. Within our framework, we employ parametric models to estimate poses for the hand and the object and employ SDF networks to learn hand and object shapes in pose-normalized coordinate frames.

$\bullet$ We show the advantage of our method by conducting comprehensive ablation experiments on ObMan. Our method produces more detailed joint reconstruction results and achieves state-of-the-art accuracy on the ObMan and DexYCB benchmarks.
\section{Related Work}
Our work focuses on joint reconstruction of hands and manipulated objects from monocular RGB images.
In this section, we first review recent methods for object shape modeling and 3D hand reconstruction.
Then, we focus on hand-and-object interaction modeling from a single color image.

\noindent \textbf{3D object modeling.}
Modeling the pose and shape of 3D objects from monocular images is one of the longest standing objectives of computer vision \cite{roberts_thesis_1963,mundy2006geometric}.
Recent methods train deep neural network models to compute the object shape~\cite{groueix2018papier,chen2019learning,mescheder2019occupancy,saito2019pifu,Xu2019disn} and pose~\cite{Li2018-fp_deepim,xiang2018posecnn,labbe2020cosypose,labbe2021robopose} directly from image pixels.
Learned object shape reconstruction from single view images has initially focused on point-cloud~\cite{qi2017pointnet}, mesh ~\cite{groueix2018papier,wang2018pixel2mesh} and voxel~\cite{choy20163d,Riegler2017OctNet} representations. 
In recent years, deep implicit representations~\cite{chen2019learning,park2019deepsdf,mescheder2019occupancy} have gained popularity.
Unlike other commonly used representations, implicit functions can theoretically model surfaces at unlimited resolution, 
which makes them an ideal choice to model detailed interactions.
We propose to leverage the flexibility of implicit functions to reconstruct hands and arbitrary unknown objects. By conditioning the signed distance function (SDF) on predicted poses, we can leverage strong shape priors from available models. 
Recent work~\cite{su2021nerf} also reveals that it is effective to encode structured information to improve the quality of NeRF~\cite{mildenhall2020nerf} for articulated bodies.

\noindent \textbf{3D hand reconstruction.}
The topic of 3D hand reconstruction has attracted wide attention since the 90s~\cite{rehg1994visual,heap1996towards}.
In the deep learning era, we have witnessed significant progress in hand reconstruction from color images. %
Most works focus on predicting 3D positions of sparse keypoints~\cite{Zimmermann2017hand3d,iqbal2018hand,sun2018integral,moon2018v2v,xiong2019a2j,mueller2018ganerated}.
These methods can achieve high accuracy by predicting each hand joint locations independently.
However sparse representations of the hand %
are insufficient to reason precisely about hand-object interactions,
which requires millimeter level accuracy.
To address this limitation, several recent works model the dense hand surface~\cite{cai20203d,kulon2019rec,Kulon2020weaklysupervisedmh,baek2019pushing,boukhayma20193d,zhou2020monocular,chen2021camera,Panteleris2018howild,mueller2019real,wang2020rgb2hands}.
A popular line of work reconstructs the hand surface by estimating
the parameters of MANO~\cite{MANO:SIGGRAPHASIA:2017},
a deformable hand mesh model.
These methods can produce anthropomorphically plausible hand meshes using
the strong hand prior captured by the parametric model.
Such methods either learn to directly regress hand mesh parameters
from RGB images~\cite{baek2019pushing,boukhayma20193d,zhou2020monocular,chen2021camera}
or fit them to a set of constraints as a post-processing step~\cite{Panteleris2018howild,mueller2019real,wang2020rgb2hands}.
Unlike previous methods, we condition the hand implicit representation on MANO parameters and produce hand reconstructions of improved visual quality.

\noindent \textbf{3D hand-object reconstruction.}
Joint reconstruction of hands and objects from monocular RGB images is a very challenging task given the partial visibility %
and strong mutual occlusions.
Methods often rely on multi-view images~\cite{oikonomidis2011full,ballan2012motion,wang2013video,hampali2020honnotate} or additional depth information~\cite{hamer2009tracking,hamer2010object,tzionas20153d,sridhar2016real,tsoli2018joint} to solve this problem.
Recent learning-based methods focus on reconstructing hands and objects directly from single-view RGB images.
To simplify the reconstruction task, several methods~\cite{hasson2020leveraging,tekin2019h+,doosti2020hopenet,yang2021cpf} make a strong assumption that the ground-truth object model is known at test-time and predict its 6D pose. Some methods propose to model hand interactions with unseen objects at test time ~\cite{hasson2019learning,romero2010b,kokic2019hho}.
Most related to our approach, Hasson~\emph{et al.}~\cite{hasson2019learning} propose a two-branch network to reconstruct the hand and an unknown manipulated object. %
The object branch uses AtlasNet~\cite{groueix2018papier} to reconstruct the object mesh and estimate its position relative to the hand.
Their method can only reconstruct simple objects which can be obtained by deforming a sphere. In contrast, SDF allows us to model arbitrary object shapes.

In order to improve the quality of hand-object reconstructions,
\cite{hasson2019learning} introduce heuristic interaction penalties at train time, Yang~\emph{et al.}~\cite{yang2021cpf} model each hand-object contact as a spring-mass system and refine the reconstruction result by an optimization process. Recent work~\cite{li2021artiboost} also applies an online data augmentation strategy to boost the joint reconstruction accuracy.
Though these methods based on parametric mesh models can achieve relatively robust reconstruction results, the modeling accuracy is limited by the underlying parametric mesh.
Closest to our approach, Karunratanakul~\emph{et al.}~\cite{karunratanakul2020grasping} propose to model the hand, the object and their contact areas using deep signed distance functions.
Their method can reconstruct hand and object meshes at a high resolution and capture detailed interactions.
However, their method is model-free and does not benefit from any prior knowledge about hands or objects. A concurrent work~\cite{ye2022s} uses an off-the-shelf hand pose estimator and leverages hand poses to improve hand-held object shapes, which operates in a less-challenging setting than ours. Different from previous works, our method brings together the advantages of both parametric models and deep implicit functions.
By embedding prior knowledge into SDFs learning, our method can produce more robust and detailed reconstruction results.

\section{Method}

\begin{figure*}[t]
  \centering
  \input{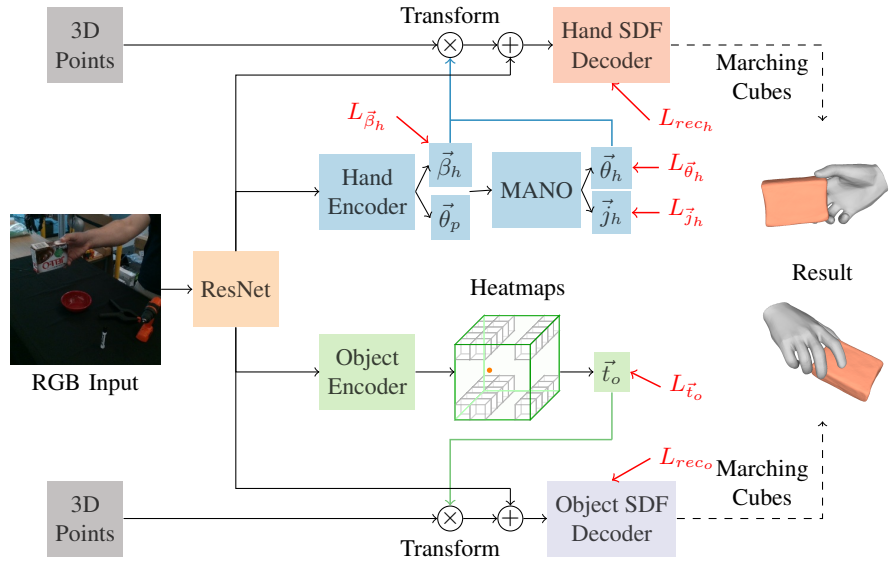}
  \caption{Our method can reconstruct detailed hand meshes and object meshes from monocular RGB images. Two gray blocks of 3D points indicate the same set of 3D query points. The red arrows denote different loss functions applied during training. The dashed arrows denote Marching Cubes algorithm~\cite{lorensen1987marching} used at test time.}
  \label{method}
\end{figure*}

As illustrated in Figure~\ref{method}, our method is designed to reconstruct the hand and object meshes from a single RGB image. Our model can be generally split into two parts: the hand part and the object part. The hand part estimates MANO parameters and uses them to transform 3D points to the hand canonical coordinate frame. Then, the hand SDF decoder predicts the signed distance for each input 3D  point and uses the Marching Cubes algorithm~\cite{lorensen1987marching} to reconstruct the hand mesh at test time. Similarly, the object part estimates the object translation relative to the hand wrist and uses it to transform the same set of 3D points. The object SDF decoder takes the transformed 3D points as input and reconstructs the object mesh. In the following, we describe the three main components of our model: hand pose estimation in Section~\ref{hpe}, object pose estimation in Section~\ref{ope}, and hand and object shape reconstruction in~Section~\ref{recon}.

\subsection{Hand pose estimation}
\label{hpe}
To embed more prior knowledge about human hands into our model, following previous works~\cite{hasson2019learning,hasson2020leveraging,yang2021cpf}, we employ a parametric hand mesh model, MANO~\cite{MANO:SIGGRAPHASIA:2017}, to capture the kinematics for the human hand. MANO is a statistical model, which could map pose ($\Vec{\theta}_{p}$) and shape ($\Vec{\beta}_{h}$) parameters to a hand mesh. To estimate hand poses, we first feed features extracted from ResNet-18~\cite{he2016deep} to the hand encoder network. The hand encoder network consists of fully connected layers and regresses $\Vec{\theta}_{p}$ and $\Vec{\beta}_{h}$. Then, we integrate MANO as a differentiable layer into our model and use it to predict the hand vertices ($\Vec{v}_{h}$), the hand joints ($\Vec{j}_{h}$) and hand poses ($\Vec{\theta}_{h}$).

We define the supervision on the joint locations ($L_{\Vec{j}_{h}}$), the shape parameters ($L_{\Vec{\beta}_{h}}$) and the predicted hand poses ($L_{\Vec{\theta}_{h}}$). To compute $L_{\Vec{j}_{h}}$, we apply L2 loss between predicted hand joints and the ground truth. However, using $L_{\Vec{j}_{h}}$ alone can result in extreme mesh deformations~\cite{hasson2019learning}. Therefore, we use another two regularization terms: $L_{\Vec{\beta}_{h}}$ and $L_{\Vec{\theta}_{h}}$. The shape regularization term ($L_{\Vec{\beta}_{h}}$) constrains that the predicted hand shape ($\Vec{\beta}_{h} \in \mathbb{R}^{10}$) is close to the mean shape in the MANO training set. The predicted hand poses ($\Vec{\theta}_{h} \in \mathbb{R}^{48}$) consist of axis-angle rotation representations for sixteen joints, including one global rotation for the wrist joint and fifteen rotations for the other local joints. The pose regularization term ($L_{\Vec{\theta}_{h}}$) constrains local joint rotations to be close to the mean pose in the MANO training set. We also apply L2 loss for the two regularization terms. For the task of hand pose estimation, the overall loss $L_{hand}$ is the summation of all $L_{\Vec{j}_{h}}$, $L_{\Vec{\beta}_{h}}$ and $L_{\Vec{\theta}_{h}}$ terms:
\begin{equation}
\begin{split}
L_{hand} = \lambda_{\Vec{j}_{h}}L_{\Vec{j}_{h}} + \lambda_{\Vec{\beta}_{h}}L_{\Vec{\beta}_{h}} + \lambda_{\Vec{\theta}_{h}}L_{\Vec{\theta}_{h}},
\end{split}
\label{loss_hand}
\end{equation}
where we set $\lambda_{\Vec{j}_{h}}$, $\lambda_{\Vec{\beta}_{h}}$ and $\lambda_{\Vec{\theta}_{h}}$ to $5\times10^{-1}$, $5\times10^{-7}$ and $5\times10^{-5}$, respectively.

\subsection{Object pose estimation}
\label{ope}
In our method, we set the origin of our coordinate system as the wrist joint defined in MANO. To solve the task of object pose estimation, we usually need to predict the object rotation and its translation. However, estimating the 3D rotation for unknown objects is a challenging and ambiguous task, especially for symmetric objects. Therefore, we here only predict the 3D object translation relative to the hand wrist. To estimate the relative 3D translation ($\Vec{t}_{o}$), we employ volumetric heatmaps~\cite{moon2018v2v,pavlakos2017coarse} to predict per voxel likelihood for the object centroid and use a soft-argmax operator~\cite{sun2018integral} to extract the 3D coordinate from heatmaps. Then, we convert the 3D coordinate into our wrist-relative coordinate system using camera intrinsics and the wrist location.

During training, we optimize network parameters by minimizing the L2 loss between the estimated 3D object translations $\Vec{t}_{o}$ and corresponding ground truth. For the task of object pose estimation, the resulting loss $L_{obj}$ is the summation of $L_{\Vec{t}_{o}}$:
\begin{equation}
\begin{split}
L_{obj} = \lambda_{\Vec{t}_{o}}L_{\Vec{t}_{o}},
\end{split}
\label{loss_obj}
\end{equation}
where we empirically set $\lambda_{\Vec{t}_{o}}$ to $5\times10^{-1}$.

\subsection{Hand and object shape reconstruction}
\label{recon}
Following previous works~\cite{park2019deepsdf,karunratanakul2020grasping}, we use neural networks to approximate signed distance functions for the hand and the object. For any input 3D point $\Vec{x}$, we employ the hand SDF decoder and the object SDF decoder to predict its signed distance to the hand surface and the object surface, respectively. However, it is very challenging to directly learn neural implicit representations for this task, because SDF networks have to handle a wide range of objects and different types of grasps. As a result, Grasping Field~\cite{karunratanakul2020grasping} cannot achieve satisfactory results in producing detailed hand-and-object interactions.

To reduce the difficulty for this task, our method makes an attempt to disentangle the shape learning and the pose learning, which could help liberate the power of SDF networks.
By estimating the hand pose, we could obtain the global rotation ($\Vec{\theta}_{hr}$) and its rotation center ($\Vec{t}_{h}$) defined by MANO. The global rotations center ($\Vec{t}_{h}$) depends on the estimated MANO shape parameters ($\Vec{\beta}_{h}$). Using the estimated $\Vec{\theta}_{hr}$ and $\Vec{t}_{h}$, we transform $\Vec{x}$ to the canonical hand pose (\emph{i.e.,} the global rotation equals to zero):
\begin{equation}
\begin{split}
\Vec{x}_{hc} = {\rm exp}(\Vec{\theta}_{hr})^{-1}(\Vec{x} - \Vec{t}_{h}) + \Vec{t}_{h},
\end{split}
\label{hand_trans}
\end{equation}
where ${\rm exp}(\cdot)$ denotes the transformation from the axis-angle representation to the rotation matrix using the \emph{Rodrigues formula}. Then, we concatenate $\Vec{x}$ and $\Vec{x}_{hc}$ and feed them to the hand SDF decoder and predict its signed distance to the hand:
\begin{equation}
\begin{split}
{\rm SDF}_{h}(\Vec{x}) = f_{h}(\Vec{I}, [\Vec{x},~\Vec{x}_{hc}]),
\end{split}
\label{handsdf}
\end{equation}
where $f_{h}$ denotes the hand SDF decoder and $\Vec{I}$ denotes image features extracted from the ResNet backbone. Benefiting from this formulation, the hand SDF encoder is aware of $\Vec{x}$ in the canonical hand pose and can focus on learning the hand shape. Similarly, by estimating the object pose, we obtain the object translation $\Vec{t}_{o}$ and transform $\Vec{x}$ to the canonical object pose:
\begin{equation}
\begin{split}
\Vec{x}_{oc} = \Vec{x} - \Vec{t}_{o}.
\end{split}
\label{obj_trans}
\end{equation}
Then, we concatenate $\Vec{x}$ and $\Vec{x}_{oc}$ and feed them to the object SDF decoder and predict its signed distance to the object:
\begin{equation}
\begin{split}
{\rm SDF}_{o}(\Vec{x}) = f_{o}(\Vec{I}, [\Vec{x},~\Vec{x}_{oc}]),
\end{split}
\label{objsdf}
\end{equation}
where $f_{o}$ denotes the object SDF decoder. By feeding $x_{oc}$ into $f_{o}$, the object SDF decoder can focus on learning the object shape in its canonical pose.

To train ${\rm SDF}_{h}(\Vec{x})$ and ${\rm SDF}_{o}(\Vec{x})$ we minimize L1 distance between predicted signed distances and corresponding ground-truth signed distances for sampled 3D points and training images. The resulting loss is the summation of $L_{rec_{h}}$ and $L_{rec_{o}}$:
\begin{equation}
\begin{split}
L_{rec} = \lambda_{rec_{h}}L_{rec_{h}} + \lambda_{rec_{o}}L_{rec_{o}},
\end{split}
\label{loss_sdf}
\end{equation}
where $L_{rec_{h}}$ and $L_{rec_{o}}$ optimize ${\rm SDF}_{h}(\Vec{x})$ and ${\rm SDF}_{o}(\Vec{x})$, respectively. We set $\lambda_{rec_{h}}$ and $\lambda_{rec_{o}}$ to $5\times10^{-1}$. In summary, we train our model in an end-to-end fashion by minimizing the sum of losses introduced above:
\begin{equation}
\begin{split}
L = L_{hand} + L_{obj} + L_{rec}.
\end{split}
\label{loss_all}
\end{equation}
Given the trained SDF networks, the hand and object surfaces are implicitly defined by the zero-level set of ${\rm SDF}_{h}(\Vec{x})$ and ${\rm SDF}_{o}(\Vec{x})$. We generate hand and object meshes using the Marching Cubes algorithm~\cite{lorensen1987marching} at test time.
\section{Experiments}
\label{exp_dex}
In this section, we present a detailed evaluation of our proposed method. We introduce benchmarks in Section~\ref{benchmarks} and describe our evaluation metrics and implementation details in Sectoins~\ref{metric}-\ref{details}. We then present hand-only ablations and hand-object experiments on the ObMan benchmark in Sections~\ref{exp_both} and~\ref{exp_hand} respectively. Finally, we present experimental results for the DexYCB benchmark  in Section~\ref{exp_dexycb}. In the appendix, we illustrate our network architecture in Section~\ref{network} and provide more implementation details in Section~\ref{train}. We also show additional qualitative results in Section~\ref{visual}.

\subsection{Benchmarks}
\label{benchmarks}
\textbf{ObMan benchmark}~\cite{hasson2019learning}.~ObMan contains synthetic images and corresponding 3D meshes for a wide range of hand-object interactions with varying hand poses and objects.
For training, we follow~\cite{park2019deepsdf,karunratanakul2020grasping} and discard meshes that contain too many double sided triangles, obtaining 87,190  samples.
For each sample, we normalize the hand mesh and the object mesh so that they fit inside a unit cube and sample 40,000 points.
At test time, we report results on 6285 samples following~\cite{karunratanakul2020grasping,hasson2019learning}.

\noindent \textbf{DexYCB benchmark}~\cite{chao2021dexycb}.~With 582K grasping frames for 20 YCB objects, DexYCB is currently the largest real benchmark for hand-object reconstruction.
Following~\cite{li2021artiboost}, we only consider right-hand samples and use the official “S0" split. We filter out the frames for which the minimum distance between the hand mesh and the object mesh is larger than 5~${\rm mm}$. We also normalize the hand mesh and the object mesh to a unit cube and sample 40,000 points to generate SDF training samples for DexYCB. As a result, we obtain 148,415 training samples and 29,466 testing samples.
\newline

\subsection{Evaluation metrics}
\label{metric}
The output of our model is structured, and a single metric does not fully capture performance. Therefore, we employ different metrics to evaluate our method. Please see Section~\ref{train} in the appendix for more details.

\noindent \textbf{Hand shape error} (${\rm {\bf{H}}_{\bf{se}}}$). We follow \cite{park2019deepsdf,karunratanakul2020grasping} and evaluate the chamfer distance between reconstructed and ground-truth hand meshes to reflect hand reconstruction accuracy.
Since the scale of the hand and the translation are ambiguous in monocular images, we optimize the scale and translation to align the reconstructed mesh with the ground-truth and sample 30,000 points from both meshes to calculate the chamfer distance. ${\rm {H}_{{se}}}$ (${\rm cm}^{2}$) is the median chamfer distance over the entire test set.

\noindent \textbf{Hand validity error} (${\rm {\bf{H}}_{\bf{ve}}}$). 
Following~\cite{Zimmermann_2019_ICCV,hampali2020honnotate} we perform \emph{Procrustes analysis} by optimizing the scale, translation and global rotation with regard to the ground-truth. We report ${\rm {{H}}_{{ve}}}$ (${\rm cm}^{2}$), the chamfer distance after alignment.

\noindent \textbf{Object shape error} (${\rm {\bf{O}}_{\bf{se}}}$). We reuse the optimized hand scale and translation from the computation of ${\rm {{H}}_{{se}}}$ to transform the reconstructed object mesh, following~\cite{karunratanakul2020grasping}. We follow the same process described for ${\rm {{H}}_{{se}}}$ to compute ${\rm {{O}}_{{se}}}$ (${\rm cm}^{2}$).

\noindent \textbf{Hand joint error} (${\rm {\bf{H}_{\bf{je}}}}$). To measure the hand pose accuracy, we compute the mean join error (${\rm cm}$) relative to the hand wrist joint over 21 joints following~\cite{Zimmermann2017hand3d}.

\noindent \textbf{Object translation error} (${\rm {\bf{O}_{\bf{te}}}}$). As we mention in Section~\ref{ope}, we predict the position of the object centroid relative to the hand wrist. We compute the L2 distance (${\rm cm}$) between the estimated object centroid and its ground-truth to report ${\rm {{O}_{{te}}}}$.

\noindent \textbf{Contact ratio} (${\rm {\bf{C}}}_{\bf{r}}$). Following~\cite{karunratanakul2020grasping}, we report the ratio of samples for which the interpenetration depth between the hand and the object is larger than zero.

\noindent \textbf{Penetration depth}. (${\rm {\bf{P}}}_{\bf{d}}$). We compute the maximum of the distances (${\rm cm}$) from the hand mesh vertices to the object's surface similarly to ~\cite{hasson2019learning,karunratanakul2020grasping}.

\noindent \textbf{Intersection volume} (${\rm {\bf{I}}}_{\bf{v}}$). Following~\cite{hasson2019learning}, we voxelize the hand and the object using a voxel size of 0.5 ${\rm cm}$ and compute their intersection volume (${\rm cm}^{3}$).

\begin{figure*}[t]
  \centering
  \input{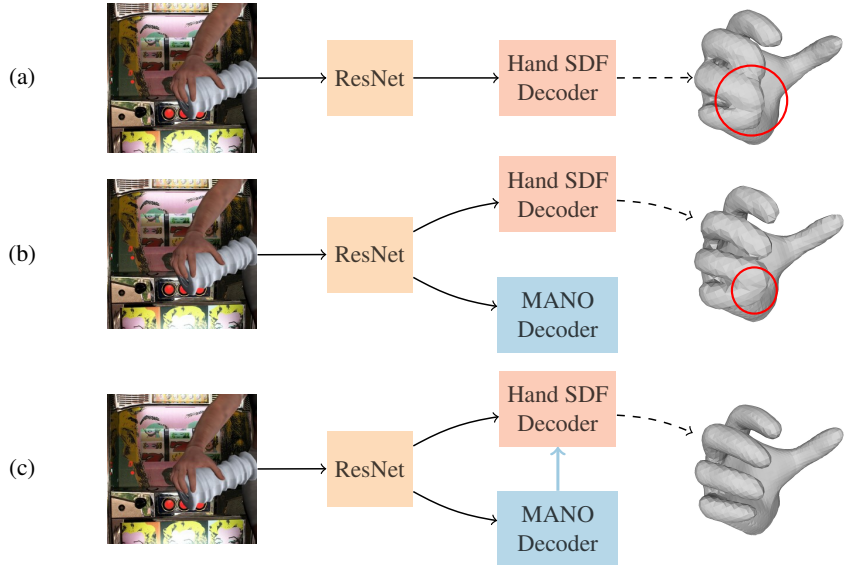}
  \caption{Three baseline models for hand-only ablation experiments. Dashed arrows denote the Marching Cubes algorithm~\cite{lorensen1987marching} used at test time.}
  \label{hand_ablation}
\end{figure*}

\subsection{Implementation details}
\label{details}
We use ResNet-18~\cite{he2016deep} as a backbone to extract features from input images of size $256 \times 256$. To construct volumetric heatmaps, we employ three deconvolution layers to consecutively upsample feature maps from $8\times8$ to $64\times64$ and set the resolution of volumetric heatmaps to $64\times64\times64$. Please see the network architecture of the SDF decoder in Section~\ref{network} in the appendix. To train hand and object SDF decoders, we randomly sample 1,000 3D points (500 positive points outside the shape and 500 negative points inside the shape) for the hand and the object, respectively. We detail our data augmentation strategies used during training in Section~\ref{train} in the appendix. We train our model with the Adam optimizer~\cite{kingma2014adam} with a batch size of 256. We set the initial learning rate to $1\times10^{-4}$ and decay it by half every 600 epoch on ObMan and every 300 epoch on DexYCB. The total number of training epochs is 1600 for ObMan and 800 for DexYCB, which takes about 90 hours on four NVIDIA 1080 Ti GPUs.

\begin{table}[t]
	\begin{minipage}{0.48\linewidth}
		\centering
		\caption{Hand-only ablation experiments using 87K ObMan training samples. }
		\setlength{\tabcolsep}{10pt}
		\begin{tabular}[t]{c|ccc}
		\hline
		Models   & ${\rm {H}_{{se}}}\downarrow$ & ${\rm {H}_{{ve}}}\downarrow$ & ${\rm {H}_{{je}}}\downarrow$  \\
		\hline
		(a)  & 0.128 & 0.113 & -\\
		(b)  & 0.126 & 0.112 & \bf{1.18}\\ 
		(c)  & \bf{0.124} & \bf{0.109} & 1.20\\
	    (${\rm c^{*}}$) & 0.101 & 0.087 & -\\
		\hline
		\end{tabular}
		\label{hand_87K}
	\end{minipage}
	\quad
	\begin{minipage}{0.48\linewidth}  
		\centering
		\centering
		\caption{Hand-only ablation experiments using 30K ObMan training samples.}
		\setlength{\tabcolsep}{10pt}
		\begin{tabular}[t]{c|ccc}
		\hline
		Models   & ${\rm {H}_{{se}}}\downarrow$ & ${\rm {H}_{{ve}}}\downarrow$ & ${\rm {H}_{{je}}}\downarrow$  \\
		\hline
		(a)  & 0.183 & 0.160 & -\\
		(b)  & 0.176 & 0.156 & \bf{1.23}\\ 
		(c)  & \bf{0.168} & \bf{0.147} & 1.27\\
		(${\rm c^{*}}$)  &  0.142&  0.126& -\\
		\hline
		\end{tabular}
		\label{hand_30K}
	\end{minipage}
\end{table}

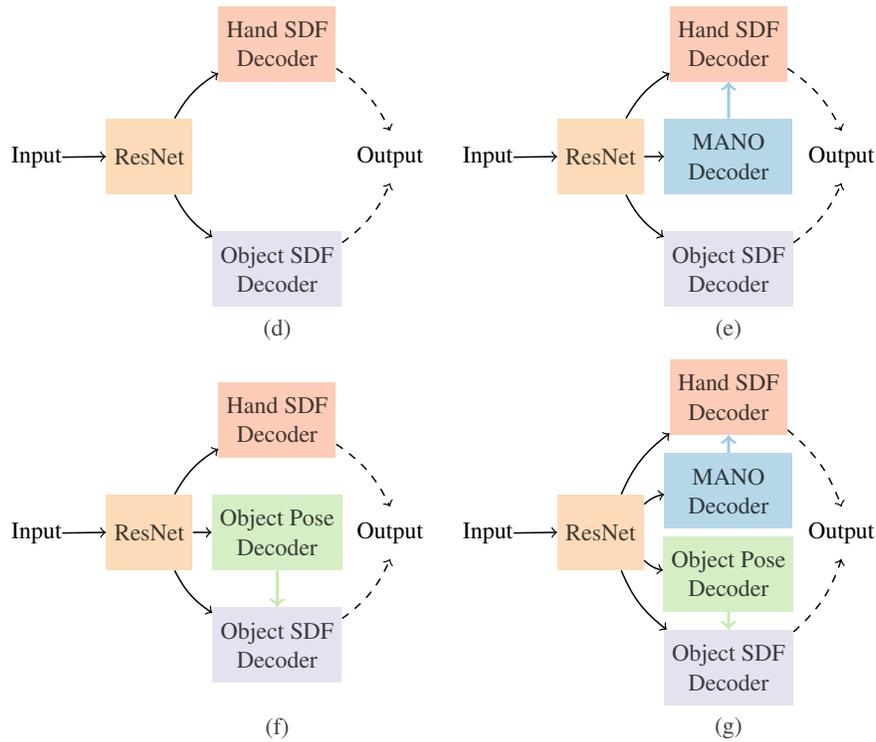
\begin{figure*}[!t]
  \centering
  \usetikzlibrary{shapes}
\newcommand{\xoffset}{6.0}
\newcommand{\yoffset}{-5.0}
\newcommand{\XD}{1.0}
\definecolor{backbone}{RGB}{253,208,162}
\definecolor{object}{RGB}{199, 233, 180}
\definecolor{hand}{RGB}{158, 202, 225}
\definecolor{handsdf}{RGB}{252, 187, 161}
\definecolor{objsdf}{RGB}{218, 218, 235}
\definecolor{query}{RGB}{175, 171, 171}
\definecolor{trans_hand}{RGB}{43,140,190}
\definecolor{trans_obj}{RGB}{116,196,118}
\begin{tikzpicture}
\node[anchor=base,inner sep=0] (ainput) at (0,-0.08){Input};
\node[rectangle,fill=backbone,minimum size=\XD cm,opacity=0.75] (aresnet) at (1.5,0) {ResNet};
\node[rectangle,fill=handsdf,minimum size=\XD cm,opacity=0.75,align=center] (ahsdf) at (3.2,1.5) {Hand SDF \\ Decoder};
\node[rectangle,fill=objsdf,minimum size=\XD cm,opacity=0.75,align=center] (aosdf) at (3.2,-1.5) {Object SDF \\ Decoder};
\node[rectangle,opacity=0.75,align=center] (caption) at (3.2,-2.3) {(d)};
\node[anchor=base,inner sep=2mm] (aoutput) at (4.7,-0.08){Output};
\draw[black,->,line width=0.2mm] (ainput.east) -- (aresnet.west);
\draw[black,->,line width=0.2mm] (aresnet) edge[bend left=15] (ahsdf);
\draw[black,->,line width=0.2mm] (aresnet) edge[bend right=15] (aosdf);
\draw[black,->,line width=0.2mm,dashed] (ahsdf) edge[bend left=15] (aoutput.north);
\draw[black,->,line width=0.2mm,dashed] (aosdf) edge[bend right=15] (aoutput.south);

\node[anchor=base,inner sep=0] (binput) at (\xoffset,-0.08){Input};
\node[rectangle,fill=backbone,minimum size=\XD cm,opacity=0.75] (bresnet) at (\xoffset+1.5,0) {ResNet};
\node[rectangle,fill=hand,minimum height=\XD cm,minimum width=1.7cm,opacity=0.75,align=center] (bmano) at (\xoffset+3.2,0.0) {MANO\\Decoder};
\node[rectangle,fill=handsdf,minimum size=\XD cm,opacity=0.75,align=center] (bhsdf) at (3.2+\xoffset,1.5) {Hand SDF \\ Decoder};
\node[rectangle,fill=objsdf,minimum size=\XD cm,opacity=0.75,align=center] (bosdf) at (3.2+\xoffset,-1.5) {Object SDF \\ Decoder};
\node[rectangle,opacity=0.75,align=center] (caption) at (3.2+\xoffset,-2.3) {(e)};
\node[anchor=base,inner sep=2mm] (boutput) at (4.7+\xoffset,-0.08){Output};
\draw[black,->,line width=0.2mm] (binput.east) -- (bresnet.west);
\draw[black,->,line width=0.2mm] (bresnet.east) -- (bmano.west);
\draw[hand,->,line width=0.4mm] (bmano.north) -- (bhsdf.south);
\draw[black,->,line width=0.2mm] (bresnet) edge[bend left=15] (bhsdf);
\draw[black,->,line width=0.2mm] (bresnet) edge[bend right=15] (bosdf);
\draw[black,->,line width=0.2mm,dashed] (bhsdf) edge[bend left=15] (boutput.north);
\draw[black,->,line width=0.2mm,dashed] (bosdf) edge[bend right=15] (boutput.south);

\node[anchor=base,inner sep=0] (cinput) at (0,-0.08+\yoffset){Input};
\node[rectangle,fill=backbone,minimum size=\XD cm,opacity=0.75] (cresnet) at (1.5,\yoffset) {ResNet};
\node[rectangle,fill=object,minimum size=\XD cm,opacity=0.75,align=center] (cobj) at (3.2,\yoffset) {Object Pose \\ Decoder};
\node[rectangle,fill=handsdf,minimum size=\XD cm,opacity=0.75,align=center] (chsdf) at (3.2,1.5+\yoffset) {Hand SDF \\ Decoder};
\node[rectangle,fill=objsdf,minimum size=\XD cm,opacity=0.75,align=center] (cosdf) at (3.2,-1.5+\yoffset) {Object SDF \\ Decoder};
\node[rectangle,opacity=0.75,align=center] (caption) at (3.2,-2.6+\yoffset) {(f)};
\node[anchor=base,inner sep=2mm] (coutput) at (4.7,-0.08+\yoffset){Output};
\draw[black,->,line width=0.2mm] (cinput.east) -- (cresnet.west);
\draw[black,->,line width=0.2mm] (cresnet.east) -- (cobj.west);
\draw[object,->,line width=0.4mm] (cobj.south) -- (cosdf.north);
\draw[black,->,line width=0.2mm] (cresnet) edge[bend left=15] (chsdf);
\draw[black,->,line width=0.2mm] (cresnet) edge[bend right=15] (cosdf);
\draw[black,->,line width=0.2mm,dashed] (chsdf) edge[bend left=15] (coutput.north);
\draw[black,->,line width=0.2mm,dashed] (cosdf) edge[bend right=15] (coutput.south);

\node[anchor=base,inner sep=0] (dinput) at (\xoffset,-0.08+\yoffset){Input};
\node[rectangle,fill=backbone,minimum size=\XD cm,opacity=0.75] (dresnet) at (1.5+\xoffset,\yoffset) {ResNet};
\node[rectangle,fill=object,minimum size=\XD cm,opacity=0.75,align=center] (dobj) at (3.2+\xoffset,\yoffset-0.55) {Object Pose \\ Decoder};
\node[rectangle,fill=hand,opacity=0.75,align=center,minimum width=1.7cm,minimum height=1.0cm] (dmano) at (3.2+\xoffset,\yoffset+0.55) {MANO \\ Decoder};
\node[rectangle,fill=handsdf,minimum size=\XD cm,opacity=0.75,align=center] (dhsdf) at (3.2+\xoffset,1.8+\yoffset) {Hand SDF \\ Decoder};
\node[rectangle,fill=objsdf,minimum size=\XD cm,opacity=0.75,align=center] (dosdf) at (3.2+\xoffset,-1.8+\yoffset) {Object SDF \\ Decoder};
\node[rectangle,opacity=0.75,align=center] (caption) at (3.2+\xoffset,-2.6+\yoffset) {(g)};
\node[anchor=base,inner sep=2mm] (doutput) at (4.7+\xoffset,-0.08+\yoffset){Output};
\draw[black,->,line width=0.2mm] (dinput.east) -- (dresnet.west);
\draw[object,->,line width=0.4mm] (dobj.south) -- (dosdf.north);
\draw[hand,->,line width=0.4mm] (dmano.north) -- (dhsdf.south);
\draw[black,->,line width=0.2mm] (dresnet) edge[bend left=15] (dmano);
\draw[black,->,line width=0.2mm] (dresnet) edge[bend right=15] (dobj);
\draw[black,->,line width=0.2mm] (dresnet) edge[bend left=15] (dhsdf);
\draw[black,->,line width=0.2mm] (dresnet) edge[bend right=15] (dosdf);
\draw[black,->,line width=0.2mm,dashed] (dhsdf) edge[bend left=15] (doutput.north);
\draw[black,->,line width=0.2mm,dashed] (dosdf) edge[bend right=15] (doutput.south);
\end{tikzpicture}
  \caption{Four models for hand-object ablation experiments. Dashed arrows denote the Marching Cubes algorithm~\cite{lorensen1987marching} used at test time.}
  \label{ho_ablation}
\end{figure*}

\subsection{Hand-only experiments on ObMan}
\label{exp_hand}
\begin{figure*}[t]
  \centering
  \input{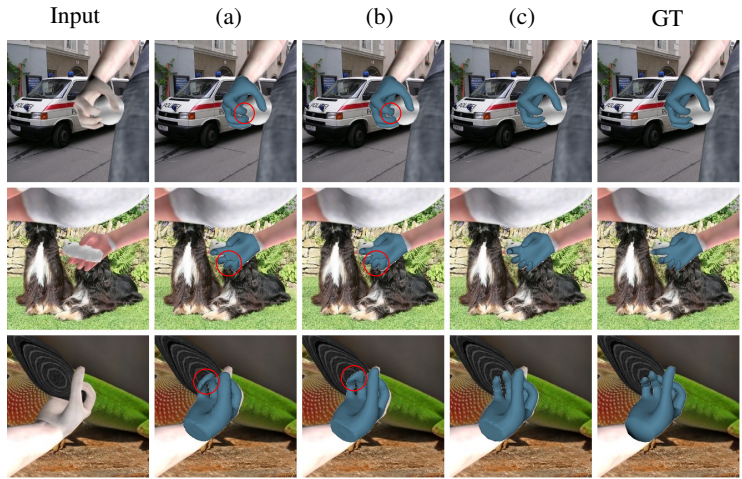}
  \caption{Qualitative comparison of hand reconstructions between different hand-only baseline models on ObMan (87K training samples).}
  \label{hand_ablation_vis} 
\end{figure*}

To validate the effectiveness of our method, we first conduct hand-only ablation experiments on ObMan. To this end, as shown in Figure~\ref{hand_ablation}, we first build three types of baseline models.
The baseline model (a) directly employs the hand SDF decoder to learn ${\rm SDF}_{h}(\Vec{x})$ from backbone features, which often results in a blurred reconstructed hand.
The baseline model (b) trains the hand SDF decoder and the MANO network jointly and achieves better results.
However, the reconstructed hand still suffers from ill-delimited outlines, which typically result in finger merging issues, illustrated in the second and third columns of Figure~\ref{hand_ablation_vis}.
Compared with the baseline model (b), the baseline model (c) further uses the estimated MANO parameters to transform sampled 3D points into the canonical hand pose, which helps disentangle the hand shape learning from the hand pose learning. As result, the hand SDF decoder can focus on learning the geometry of the hand and reconstruct a clear hand. The model (${\rm c^{*}}$) uses ground-truth hand poses, which is the upper-bound of our method. Tables~\ref{hand_87K} and~\ref{hand_30K} present quantitative results for these four models. In Table~\ref{hand_87K}, we present our results using all ObMan training samples and observe that the baseline model (c) has the lowest ${\rm {H}_{{se}}}$ and ${\rm {H}_{{ve}}}$, which indicates that it achieves the best hand reconstruction quality. The baseline model (c) can also perform hand pose estimation well and reduce the joint error to 1.2 ${\rm cm}$.
It shows that the model (c) can transform $\Vec{x}$ to the hand canonical pose well with reliable $\Vec{\theta}_{hr}$ and $\Vec{t}_{h}$ and benefit the learning of the hand SDFs.
In Figure~\ref{hand_ablation_vis}, we also visualize results obtained from different models and observe that our method can produce more precise hands even under occlusions. To check whether our method can still function well when the training data is limited, we randomly choose 30K samples to train these three models and summarize our results in Table~\ref{hand_30K}. We observe that the advantage of the model (c) is more obvious using less training data. When compared with the model (a), our method can achieve more than 8\% improvement in ${\rm {H}_{{se}}}$ and ${\rm {H}_{{ve}}}$, which further validates the effectiveness of our approach.

\begin{table}[t]
\centering
\caption{Hand-object ablation experiments using 87K ObMan training data.}
\setlength{\tabcolsep}{8pt}
\begin{tabular}{c|cccccccc}
\hline
Models & ${\rm {H}_{{se}}}\downarrow$ & ${\rm {H}_{{ve}}}\downarrow$ & ${\rm {O}_{{se}}}\downarrow$ & ${\rm {H}_{{je}}}\downarrow$ & ${\rm {O}_{{te}}}\downarrow$ & ${\rm {C}_{{r}}}$ & ${\rm {P}_{{d}}}$ & ${\rm {I}_{{v}}}$ \\ \hline
(d)&0.140&0.124&4.09&-&-&90.3\%&0.50&1.51               \\ 
(e)&\bf{0.131}&\bf{0.114}&4.14&\bf{1.12}&-&94.7\%&0.58&2.00               \\ 
(f)&0.148&0.130&\bf{3.36}&-&\bf{3.29}&92.5\%&0.57&2.26               \\ 
(g)&0.136&0.121&3.38&1.27&3.29&95.5\%&0.66&2.81\\
(${\rm g^{*}}$)&0.111&0.093&2.11&-&-&94.5\%&0.76&3.87\\ \hline
\end{tabular}
\label{ho_ablation_exp}
\end{table}

\begin{table}[!t]
\centering
\caption{Comparison with previous state-of-the-art methods on ObMan.}
\setlength{\tabcolsep}{4pt}
\begin{tabular}{c|cccccccc}
\hline
Methods & ${\rm {H}_{{se}}}\downarrow$ & ${\rm {H}_{{ve}}}\downarrow$ & ${\rm {O}_{{se}}}\downarrow$ & ${\rm {H}_{{je}}}\downarrow$ & ${\rm {O}_{{te}}}\downarrow$ & ${\rm {C}_{{r}}}$ & ${\rm {P}_{{d}}}$ & ${\rm {I}_{{v}}}$ \\ \hline
Hasson \emph{et al.}~\cite{hasson2019learning}&0.415&0.383&3.60&\bf{1.13}&-&94.8\%&1.20&6.25\\ 
Karunratanakul \emph{et al.}~\cite{karunratanakul2020grasping}-1De&0.261&0.246&6.80&-&-&5.63\%&0.00&0.00\\ 
Karunratanakul \emph{et al.}~\cite{karunratanakul2020grasping}-2De&0.237&-&5.70&-&-&69.6\%&0.23&0.20\\
Ours (g)&\bf{0.136}&\bf{0.121}&\bf{3.38}&1.27&\bf{3.29}&95.5\%&0.66&2.81               \\ \hline
\end{tabular}
\label{sota}
\end{table}

\subsection{Hand-object experiments on ObMan}
\label{exp_both}
\begin{figure*}[t]
  \centering
  \input{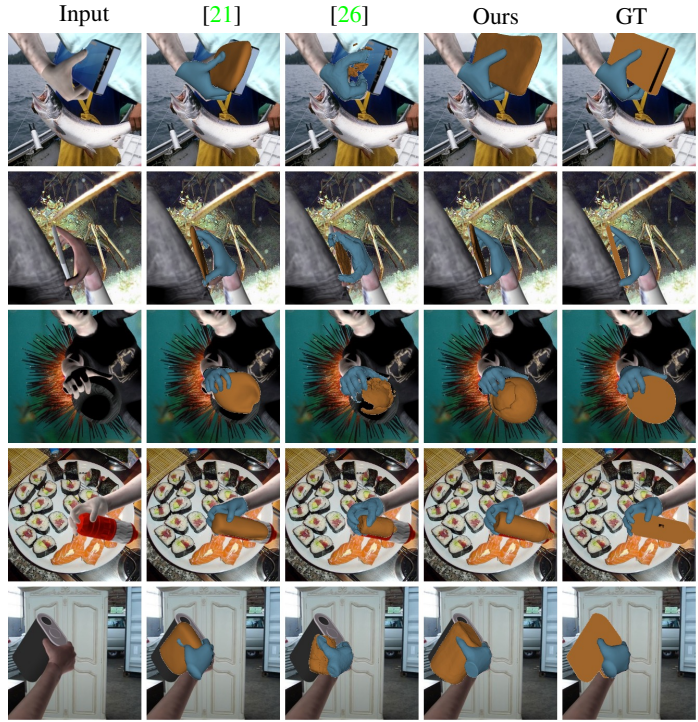}
  \caption{Qualitative comparison between different types of methods in hand-object experiments on ObMan. Compared with recent methods~\cite{hasson2019learning,karunratanakul2020grasping}, our approach produces more precise reconstructions both for the hands and objects.}
  \vspace{-0.2cm}
  \label{ho_ablation_vis}
\end{figure*}

Given promising results for hand-only experiments, we next validate our approach for the task of hand-object reconstruction. As shown in Figure~\ref{ho_ablation}, we first build four baseline models. The baseline model (d) directly uses the hand and the object decoder to learn SDFs. Compared with the model (d), the model (e) estimates MANO parameters for the hand and uses it to improve the learning the hand SDF decoder. The model (f) estimates the object pose and uses the estimated pose to learn the object SDF decoder. The model (g) combines models (e) and (f) and uses estimated hand and object poses to improve the learning of the hand SDFs and the object SDFs, respectively. The model ($\rm {g^*}$) is trained with ground-truth hand poses and object translations, which serves as the upper-bound for our method. We summarize our experimental results for these five models in Table~\ref{ho_ablation_exp}. 
Compared with the baseline model (d), the model (e) achieves a 6.4\% and 8.8\% improvement in ${\rm {H}_{{se}}}$ and ${\rm {H}_{{ve}}}$, respectively. It shows that embedding hand prior knowledge and aligning hand poses to the canonical pose can improve learning the hand SDFs. By comparing the model (f) with the baseline model (d), we align object poses to their canonical poses using estimated object pose parameters and greatly reduce ${\rm {O}_{{se}}}$ from 4.09 ${\rm cm}^{2}$ to 3.36 ${\rm cm}^{2}$. Finally, our full model (g) combines the advantages of models~(e) and~(f) and can produce high-quality hand meshes and object meshes. In Table~\ref{sota}, we compare our method against previous state-of-the-art methods and show that our approach outperforms previous state-of-the-art methods~\cite{hasson2019learning,karunratanakul2020grasping} by a significant margin. 
When we take a closer look at metrics (${\rm {C}_{{r}}}$, ${\rm {P}_{{d}}}$, ${\rm {I}_{{v}}}$) that reflect hand-object interactions, we can observe that the reconstructed hand and the reconstructed object from our model are in contact with each other in more than 95.5\% of test samples. Compared with the SDF method~\cite{karunratanakul2020grasping}, our method encourages the contact between the hand mesh and the object mesh. Compared with the MANO-based method~\cite{hasson2019learning}, the penetration depth (${\rm {P}_{{d}}}$) and intersection volume (${\rm {I}_{{v}}}$) of our model is much lower, which suggests that our method can produce more detailed hand-object interactions. In Figure~\ref{ho_ablation_vis}, we also visualize reconstruction results from different methods. Compared to previous methods, our model can produce more realistic joint reconstruction results even for objects with thin structures. We include more qualitative analysis on ObMan in Section~\ref{visual} in the appendix.

\begin{table}[!t]
\centering
\caption{Comparison with previous state-of-the-art methods on DexYCB.}
\setlength{\tabcolsep}{4pt}
\begin{tabular}{c|cccccccc}
\hline
Method & ${\rm {H}_{{se}}}\downarrow$ & ${\rm {H}_{{ve}}}\downarrow$ & ${\rm {O}_{{se}}}\downarrow$ & ${\rm {H}_{{je}}}\downarrow$ & ${\rm {O}_{{te}}}\downarrow$ &${\rm {C}_{{r}}}$ & ${\rm {P}_{{d}}}$ & ${\rm {I}_{{v}}}$ \\ \hline
Hasson \emph{et al.}~\cite{hasson2019learning} &0.785&0.594&4.4&2.0&-&95.8\%&1.32&7.67\\
Karunratanakul \emph{et al.}~\cite{karunratanakul2020grasping} &0.741&0.532&5.8&-&-&96.7\%&0.83&1.34\\
Ours (g) &\bf{0.523}&\bf{0.375}&\bf{3.5}&\bf{1.9}&\bf{2.7}&96.1\%&0.71&3.45\\ \hline
\end{tabular}
\label{ho_dexycb_exp}
\vspace{-0.3cm}
\end{table}

\subsection{Hand-object experiments on DexYCB}
\label{exp_dexycb}
\begin{figure*}[t]
  \centering
  \input{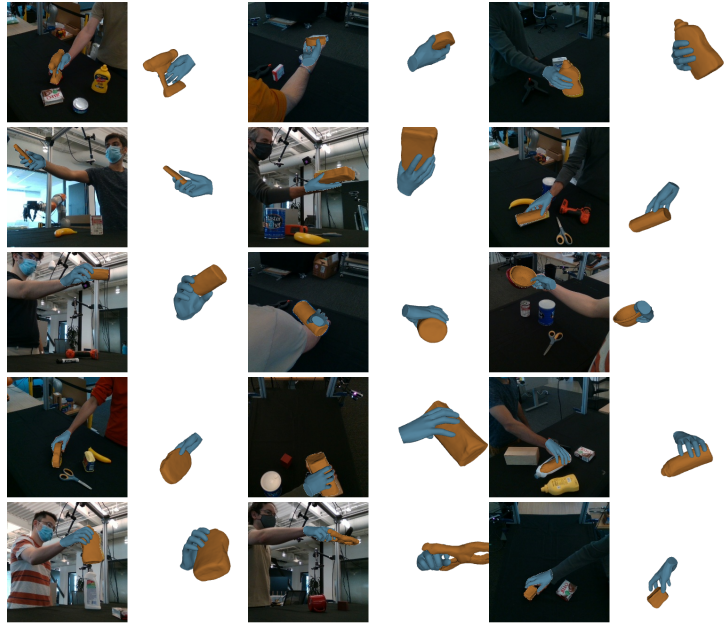}
  \caption{Qualitative results of our model on the DexYCB benchmark. Our model produces convincing 3D hand-and-object reconstruction results in the real-world setting.}
  \label{dexycb_vis}
  \vspace{-0.5cm}
\end{figure*}

To validate our method on real data, we next present experiments on the DexYCB benchmark and compare our results to the state of the art. We summarize our experimental results in Table~\ref{ho_dexycb_exp}. Compared with previous methods, we achieve a 29.4\% improvement in ${\rm {H}_{{se}}}$ and a 20.5\% improvement in ${\rm {O}_{{se}}}$, which shows that our method improves both the hand and object reconstruction accuracy. The hand-object interaction metrics for DexYCB also indicate that our method works well for real images.
Figure~\ref{dexycb_vis} illustrates qualitative results of our method on the DexYCB benchmark. We can observe that our method can accurately reconstruct hand shapes under different poses and a wide range of real-world objects. Please see more qualitative results on DexYCB in Section~\ref{visual} in the appendix.

\section{Conclusion}
In this work, we combine advantages of parametric mesh models and SDFs for the task of a joint hand-object reconstruction. To embed prior knowledge into SDFs and to increase the learning efficiency, we propose to disentangle the shape learning and pose learning for both the hand and the object. Then, we align SDF representations with respect to estimated poses and enable learning of more accurate shape estimation. Our model outperforms previous state-of-the-art methods by a significant margin on main benchmarks. Our results also demonstrate significant improvements in visual quality.

\noindent \textbf{Acknowledgements.} This work was granted access to the HPC resources of IDRIS under the allocation AD011013147 made by GENCI. This work was funded in part by the French government under management of Agence Nationale de la Recherche as part of the “Investissements d’avenir” program, reference ANR19-P3IA-0001 (PRAIRIE 3IA Institute) and by Louis Vuitton ENS Chair on Artificial Intelligence.

\clearpage

\bibliographystyle{splncs04nat}
\bibliography{egbib}
\newpage
\appendix
\clearpage
\counterwithin{figure}{section}
\counterwithin{table}{section}
\counterwithin{equation}{section}

\pagenumbering{Roman}  

\newpage
\vskip .375in
\begin{center}
{\Large \bf AlignSDF: Pose-Aligned Signed Distance Fields \\for Hand-Object Reconstruction \\ \vspace{0.5cm} \large Appendix \par}
  \vspace*{24pt}
  {
  \par
  }
\end{center}

In this appendix, we provide additional details for our experimental settings as well as qualitative results of our method. We first present details for our network architecture in Section~\ref{network}. Section~\ref{train} then provides additional implementation details for our training and evaluation procedures. Finally, we present and discuss additional qualitative results in Section~\ref{visual}.
\vspace{-0.4cm}

\section{Network Architecture}
\label{network}
\vspace{-1.2cm}

\begin{figure*}[!ht]
  \centering
  \usetikzlibrary{shapes}
\newcommand{\xd}{0.5}
\newcommand{\XD}{1.0}
\newcommand{\xoffset}{6}
\newcommand{\yoffset}{-4.8}
\definecolor{backbone}{RGB}{253,208,162}
\definecolor{obj}{RGB}{127,205,187}
\definecolor{object}{RGB}{199, 233, 180}
\definecolor{hand}{RGB}{158, 202, 225}
\definecolor{handsdf}{RGB}{252, 187, 161}
\definecolor{objsdf}{RGB}{218, 218, 235}
\definecolor{query}{RGB}{175, 171, 171}
\definecolor{trans_hand}{RGB}{43,140,190}
\definecolor{trans_obj}{RGB}{116,196,118}
\definecolor{fc}{RGB}{117,107,177}
\begin{tikzpicture}
\node[rectangle,draw=red!60,line width=0.2mm,minimum height=2.4 cm,opacity=0.75,align=center,inner sep=1em] (box) at (0,-0.4) {};

\node[rectangle,fill=query,minimum width=\xd cm,opacity=0.75,rotate=90] (point) at (0,-1.2) {6};
\node[rectangle,fill=backbone,minimum width=\XD cm,opacity=0.75,align=center,rotate=90] (img_feat) at (0,0) {256};
\node[rectangle,fill=backbone,minimum size=\xd cm,opacity=0.75,inner sep=0.5em] (resnet) at (-1.2,0) {ResNet};
\node[rectangle,minimum width=\xd cm,opacity=0.75,inner sep=0.5em] (images) at (-3, 0) {Image};
\node[rectangle,minimum width=\xd cm,opacity=0.75,inner sep=0.5em] (point_text) at (-2.5,-1.2) {Point Features};
\node[rectangle,minimum width=\xd cm,opacity=0.75,inner sep=0.5em] (point_textb) at (-2.5,-1.6) {($[\Vec{x}, \Vec{x}_{hc}]$ or $[\Vec{x}, \Vec{x}_{oc}]$)};

\node[rectangle,fill=backbone,line width=0.2mm,minimum width=2.1 cm,minimum height=0.2cm,opacity=0.75,align=center,rotate=90] (fc1) at (1.2,-0.4) {512};
\node[rectangle,fill=backbone,line width=0.2mm,minimum height=0.2 cm,minimum width=1.2cm,opacity=0.75,align=center,rotate=90] (fc2) at (2.4,-1.1) {250};
\node[rectangle,fill=backbone,line width=0.2mm,minimum height=0.2 cm,minimum width=1.2cm,opacity=0.75,align=center,rotate=90] (feat) at (2.4,0.2) {262};
\node[rectangle,draw=red!60,line width=0.2mm,minimum height=2.8 cm,opacity=0.75,align=center,inner sep=1em] (box2) at (2.4,-0.4) {};
\node[rectangle,line width=0.2mm,minimum height=0.2 cm,minimum width=1.2cm,opacity=0.75,align=center,rotate=90] (featu) at (2.4,1.4) {};
\node[rectangle,fill=backbone,line width=0.2mm,minimum width=2.1 cm,minimum height=0.2cm,opacity=0.75,align=center,rotate=90] (fc3) at (3.6,-0.4) {512};
\node[rectangle,fill=backbone,line width=0.2mm,minimum width=2.1 cm,minimum height=0.2cm,opacity=0.75,align=center,rotate=90] (fc4) at (4.8,-0.4) {512};
\node[rectangle,fill=backbone,line width=0.2mm,minimum width=0.4 cm,minimum height=0.2cm,opacity=0.75,align=center,rotate=90] (fc5) at (6,-0.4) {1};
\node[rectangle,minimum width=\xd cm,opacity=0.75,inner sep=0.5em](output_text) at (6.2, -1) {Signed Distance};

\draw[fc,line width=0.3mm,->] (box.east) -- (fc1.north);
\draw[fc,line width=0.3mm,->] (fc1.south) -- (fc2.north);
\draw[black,line width=0.2mm,-] (box.north) |- (featu.center);
\draw[black,line width=0.2mm,->] (featu.center) -- (feat.east);
\draw[fc,line width=0.3mm,->] (box2.east) -- (fc3.north);
\draw[fc,line width=0.3mm,->] (fc3.south) -- (fc4.north);
\draw[fc,line width=0.3mm,->] (fc4.south) -- (fc5.north);
\draw[black,line width=0.2mm,->] (point_text.east) -- (point.north);
\draw[black,line width=0.2mm,->] (resnet.east) -- (img_feat.north);
\draw[black,line width=0.2mm,->] (images.east) -- (resnet.west);
\end{tikzpicture}
  \caption{Network architecture used for our hand and object SDF decoders. Following~\cite{karunratanakul2020grasping}, we also use five fully connected layers (marked in purple) for the SDF decoder. The number in the box denotes the dimension of features. $\Vec{x}$ denotes the original 3D coordinate. $\Vec{x}_{hc}$ and $\Vec{x}_{oc}$ denote the transformed 3D coordinate in the hand and object canonical coordinate system, respectively.}
  \label{arch}
\end{figure*}
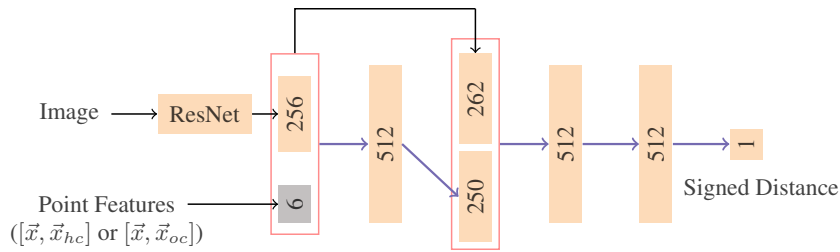

Following previous works~\cite{hasson2019learning,karunratanakul2020grasping}, we use ResNet-18~\cite{he2016deep} as our backbone network. To achieve a fair comparison with the previous method~\cite{karunratanakul2020grasping}, as shown in Figure~\ref{arch}, we also use five fully connected layers to estimate the signed distance from the query point to the hand surface or the object surface. The SDF decoder takes the 256-dimensional image features and 6-dimensional point features as inputs. The image features are extracted from the ResNet-18 backbone. Following Equation {\color{red} 3} and Equation {\color{red} 5} in our paper, we transform the original 3D point $\Vec{x}$ into its counterpart $\Vec{x}_{hc}$ in the hand canonical coordinate system or its counterpart $\Vec{x}_{oc}$ in the object canonical coordinate system. Then, we construct point features by concatenating $\Vec{x}$ and $\Vec{x}_{hc}$ for the hand SDF decoder or by concatenating $\Vec{x}$ and $\Vec{x}_{oc}$ for the object SDF decoder.

\section{Training and Evaluation}
\label{train}

We train all of our models with the following data augmentation. We randomly rotate the input image and 3D points in the camera coordinate system. We empirically find that this data augmentation can boost the performance for 3D reconstruction. We randomly augment training samples via [$-45^{\circ},45^{\circ}$] rotation for our experiments on ObMan~\cite{hasson2019learning} or [$-15^{\circ},15^{\circ}$] rotation for our experiments on DexYCB~\cite{chao2021dexycb}.

We set the hand wrist joint defined by MANO~\cite{MANO:SIGGRAPHASIA:2017} as the origin of our coordinate system. In training, we use a fixed scaling factor to scale all negative points (\emph{i.e.,} points that lie in the hand or object mesh) across the dataset within a unit cube. This results in a scaling factor of 7.02 and 6.21 on ObMan and DexYCB, respectively.

To measure the physical quality of our joint reconstruction, we report Contact Ratio~(${\rm C}_r$), Penetration Depth~(${\rm P}_d$) and Intersection Volume~(${\rm I}_v$). We use the trimesh library to detect whether there exists a collision between the hand mesh and the object mesh and compute the max penetration depth between two meshes. We follow the same process as~\cite{karunratanakul2020grasping,karunratanakul2021skeleton} to compute ${\rm I}_v$.

\section{Qualitative results}
\label{visual}

\begin{figure*}[t]
  \centering
  \input{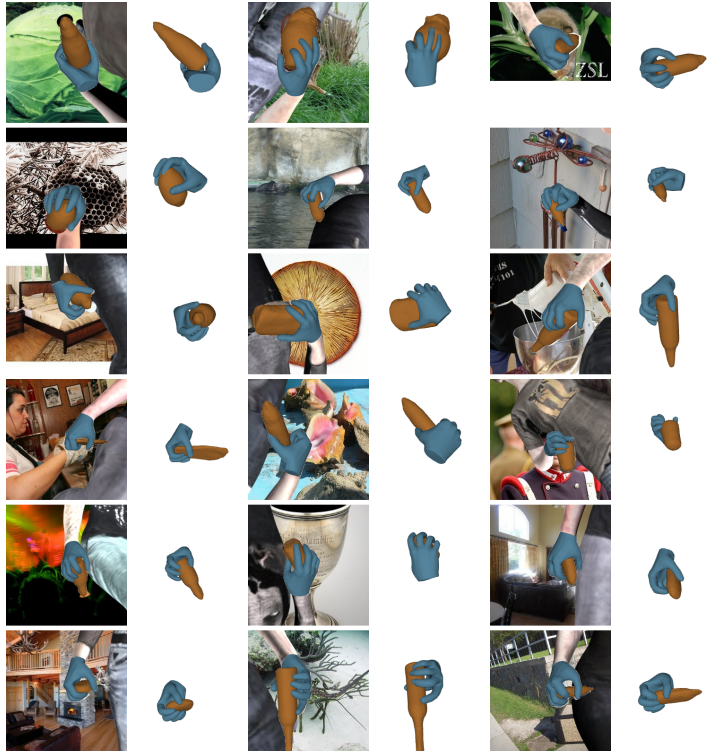}
  \caption{Qualitative results of our method on the ObMan~\cite{hasson2019learning} benchmark. Our method can produce convincing 3D reconstruction results even in cluttered scenes.}
  \label{obman_demo} 
\end{figure*}

\begin{figure*}[t]
  \centering
  \input{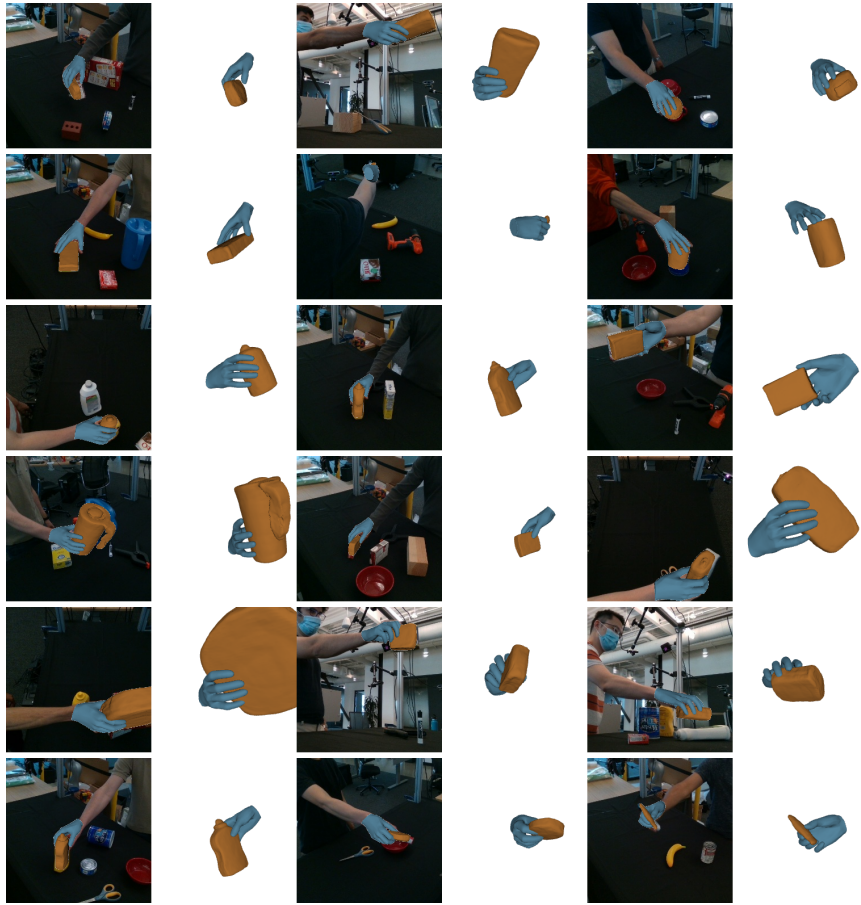}
  \caption{Qualitative results of our method on the DexYCB~\cite{chao2021dexycb} benchmark. Our method can also produce realistic 3D reconstruction results for real scenes.}
  \label{dexycb_demo} 
\end{figure*}

\begin{figure*}[t]
  \centering
  \input{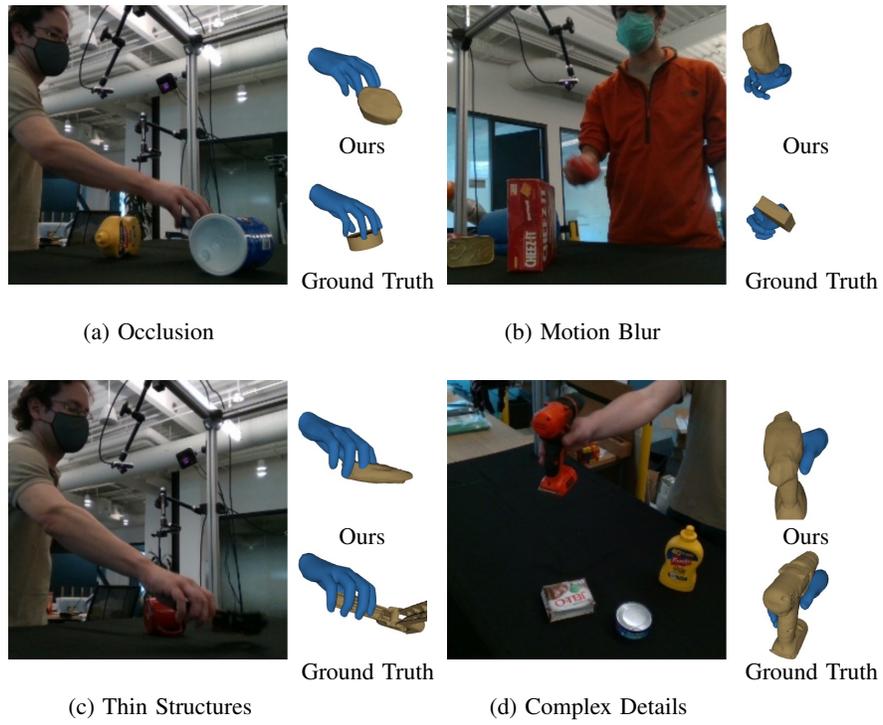}
  \caption{Failure cases of our method on the DexYCB~\cite{chao2021dexycb} benchmark.}
  \label{failure} 
\end{figure*}

We present additional qualitative results on ObMan~\cite{hasson2019learning} in Figure~\ref{obman_demo} and DexYCB~\cite{chao2021dexycb} in Figure~\ref{dexycb_demo}. We also study failure cases on DexYCB in Figure~\ref{failure}. 
From Figure~\ref{obman_demo}, we observe that our method can deal with a wide range of objects and recovers detailed interactions between the hand and the object. In Figure~\ref{dexycb_demo} we show qualitative results of our method for real images from the DexYCB benchmark. We can see that our method can reconstruct objects of different sizes and often achieve the excellent reconstruction of hands and objects. 

While our method advances the state of the art accuracy by a significant margin, it still does not achieve satisfactory performance in some cases. 
In Figure~\ref{failure} we show four typical failure cases on DexYCB. As shown in Figure~\ref{failure}(a), when the hand or the object is heavily occluded, our method sometimes cannot make robust predictions. In Figure~\ref{failure}(b), we show that motion blur in input images might also disturb 3D reconstruction results. As shown in Figure~\ref{failure}(c,~d), the recovery of thin structures and objects with complex shapes remains challenging. To deal with these issues, future works could leverage the temporary information from videos to filter input noise and gather more details about 3D scenes.
\end{document}